\documentclass[letterpaper]{article}
\usepackage{proceed2e}
\usepackage[margin=1in]{geometry}

\usepackage{times}

\usepackage{amsmath}
\usepackage{amssymb}
\usepackage{tikz}
\usepackage{pgfplots}
\usepackage{subcaption}
\pgfplotsset{compat=1.11}
\usepackage{bm}
\usetikzlibrary{positioning}
\usepackage[]{algorithm2e}

\newtheorem{theorem}{Theorem}

\title{A Forest Mixture Bound for Block-Free Parallel Inference}

\author{ {\bf Neal G. Lawton} \\
Computer Science Dept. \\
University of Southern California\\
Los Angeles, CA 90007 \\
\And
{\bf Aram  Galstyan} \\
Computer Science Dept. \\
University of Southern California\\
Los Angeles, CA 90007 \\
\And
{\bf Greg Ver Steeg} \\
Computer Science Dept. \\
University of Southern California\\
Los Angeles, CA 90007 \\
}

\begin{document}

\maketitle

\begin{abstract}

Coordinate ascent variational inference is an important algorithm for inference in probabilistic models, but it is slow because it updates only a single variable at a time.
Block coordinate methods perform inference faster by updating blocks of variables in parallel.
However, the speed and stability of these algorithms depends on how the variables are partitioned into blocks.
In this paper, we give a stable parallel algorithm for inference in deep exponential families that doesn't require the variables to be partitioned into blocks.
We achieve this by lower bounding the ELBO by a new objective we call the \textit{forest mixture bound} (FM bound) that separates the inference problem for variables within a hidden layer.
We apply this to the simple case when all random variables are Gaussian and show empirically that the algorithm converges faster for models that are inherently more forest-like.
\end{abstract}

\section{INTRODUCTION}

Inference in directed models like deep exponential families (DEF's) \cite{blei_DEF} is complicated by the ``explaining away effect'': for a directed model with observed variables $\bm x \in \mathbb R^n$ and latent variables $\bm y \in \mathbb R^m$, independent ``causes'' $y_j$ become dependent given an observed ``effect'' $x_i$. To handle this, the coordinate ascent variational inference (CAVI) algorithm iteratively updates the variational distribution for a single latent variable $y_j$ while holding the variational distribution for all other latent variables fixed \cite{blei_tutorial}.

Though the $y_j$'s are not conditionally independent given $\bm x$ except in exceedingly simple models, in many cases the $y_j$'s are \textit{nearly} conditionally independent. Is there a way to perform stable parallel inference in such models, or do we have to resort to the serial coordinate algorithm?

Block methods provide one avenue for parallel inference. These algorithms work by first partitioning the latent variables into a collection of blocks, and then iteratively updating a variable from each block in parallel. However, the speed (as in MCMC methods \cite{asynchronous_gibbs}) or stability (as in Hogwild methods \cite{hogwild_sgd}) of the resulting algorithm will depend on how the variables are blocked, and finding a good choice of blocking for an arbitrary model can be difficult.

The main contribution of this paper is a novel lower
bound on log-likelihood we call the forest mixture bound
(FM bound) that separates the problem of inference for
each variable in a hidden layer. This allows all the
variables in a layer to be updated in parallel, without the
use of blocks. We call the resulting stable parallel inference algorithm the forest mixture algorithm (FM algorithm).

We study in detail the case when all the random variables in the DEF are Gaussian. 
We then demonstrate on both synthetic and real-world data the proposed stable method achieves faster convergence compared to existing methods.

\section{RELATED WORK}
\paragraph{Hogwild Block Methods} There are two types of block methods for inference. The first is Hogwild-type algorithms \cite{hogwild_sgd}\cite{desa_2016} \cite{DBLP:journals/corr/WangB14a} \cite{NIPS2014_5614}. After partitioning the variables into blocks, these algorithms iteratively choose a single variable from each block and update as in CAVI, but in parallel \cite{desa_2016}. These algorithms are guaranteed to be stable only in certain cases, e.g., when the blocks are conditionally independent \cite{hogwild_gaussian}. 

\paragraph{Stable Block Methods} Instead of making CAVI updates in parallel, stable block algorithms achieve stability by making small parallel updates \cite{pmlr-v5-sontag09a}. For example, ``exact'' asynchronous Gibbs sampling randomly rejects each block update according to an MCMC rejection ratio \cite{asynchronous_gibbs}. If the blocks are chosen poorly, the rejection rate will increase and the rate of convergence will decrease \cite{particle_gibbs}.

In either type of block method, the performance of the algorithm depends on how the variables are blocked. In a distributed computation setting, blocking is necessary since each worker can only store a fraction of all variables in local memory. In this case, the FM bound provides a method for updating variables within a block or worker in parallel, instead of updating only a single variable in each block at a time.

\paragraph{Amortized Inference} Instead of treating inference as an inverse problem that has to be solved for each observation, VAE's train inference network (encoder) so the cost of inference is amortized over many observations \cite{vae}. Once the encoder is trained, inference for any observation can be performed quickly with a single pass through the inference network. Encoder-free methods like ours may still be useful in the case when we have a trained generative model (decoder) but no trained encoder and want to perform inference for only a few samples or, more likely, for when we want to improve the solution produced by the encoder at test time.

\paragraph{Undirected Models} Besides directed models, there is a wide literature for fast inference in undirected models \cite{baque_2016} \cite{singh_2010}. Note that inference in undirected models like Deep Restricted Boltzmann Machines \cite{hinton_2009} can already be parallelized: non-consecutive layers can be updated in parallel in red-black fashion. In fact, the same degree of parallelization can be achieved in a directed model using our technique. While there is also a wide literature on bounding the log-partition function of an \textit{undirected} model \cite{wainwright_2005}, we derive the FM bound by lower bounding the log-partition function of a \textit{directed} model. The technique we use may be applicable to undirected models, but that is not explored in this paper.

\paragraph{Structure Learning}

The FM bound we derive is closely related to an interesting family of models called \textit{forest mixture models}. These models may be applicable to the problem of structure learning, where the task is to infer the graphical structure of the underlying model from data \cite{chow_liu}. However, in this paper we narrowly focus on the problem of inference in a \textit{given} generative model, not on training a new one.

\section{PRELIMINARIES}
Vector-valued variables are written in bold. The component-wise product of two vectors $\bm u$ and $\bm v$ is denoted $u \odot v$. Unless stated otherwise, all expectations, including the variance $\text{Var}[\cdot]$, standard deviation $\text{Std}[\cdot]$, and conditional entropy $H(\bm y | \bm x)$, are taken with respect to the variational distribution $q(\bm y | \bm x)$, though we sometimes write this explicitly for emphasis.

An \textit{exponential family} of distributions is a family of distributions of the form

\begin{equation*}
p(x) = \exp \{ g(x) + t(x) \cdot \eta - a(\eta) \}
\end{equation*}

Where $g$ is the log-base measure, $t$ are the sufficient statistics, $\eta$ are the natural parameters, and $a$ is the log-partition function. When $\eta$ is a function of another random variable $\bm y$, e.g., $\eta = b + \bm w \cdot \bm y$, we will sometimes write $\eta = \eta(\bm y)$ for emphasis.

We denote the Gaussian probability density function with mean $\mu$ and variance $\sigma^2$ as $\mathcal N(\mu, \sigma^2)$. When we write \mbox{$\log p(x) \propto f(x)$}, we mean $\log p(x) = f(x) + constant$.
\subsection{FOREST MIXTURE MODELS}
\begin{figure*}[h]
	\centering
	\begin{subfigure}[t]{0.2\textwidth}
		\centering
		\begin{tikzpicture}[every node/.style={draw,circle}]
		\foreach \x in {1,...,5}
			\node (x-\x) at (\x * 0.8,0) {$x_\x$};
		\foreach \y in {1,...,3}
			\node (y-\y) at (\y+0.3333, 1.5) {$y_\y$};
		\foreach \x in {1,...,5}
			\foreach \y in {1,...,3}
				\draw [densely dotted,->] (y-\y) -- (x-\x);
		\end{tikzpicture}
		\caption{}\label{fig:fmm:random_edges}
	\end{subfigure}
	\hspace{1cm}
	\begin{subfigure}[t]{0.2\textwidth}
		\centering
		\begin{tikzpicture}[every node/.style={draw,circle}]
		\foreach \x/\r in {1/305, 2/105, 3/255, 4/15, 5/195} {
			\draw[line width=.3mm, color=red,   shift={(\x * 0.8, 0)}, rotate=0  ] (0.35,0) arc (0:120:0.35);
			\draw[line width=.3mm, color=green, shift={(\x * 0.8, 0)}, rotate=120] (0.35,0) arc (0:120:0.35);
			\draw[line width=.3mm, color=blue,  shift={(\x * 0.8, 0)}, rotate=240] (0.35,0) arc (0:120:0.35);
			\filldraw[line cap=round, line width=.01mm, shift={(\x * 0.8, 0)}, rotate=\r] (.3,0) -- (-5:.225) arc (-5:5:.225) -- cycle;
			\node (x-\x) at (\x * 0.8,0) {{$x_\x$}};
		}
		\foreach \y/\color in {1/red, 2/green, 3/blue}
			\node[color=\color, text=black] (y-\y) at (\y+0.3333, 1.5) {$y_\y$};
		\foreach \x/\y in {1/1, 1/2, 2/2, 2/3, 3/1, 3/2, 4/2, 4/3, 5/1, 5/3}
			\draw [densely dotted, color=gray] (y-\y) -- (x-\x);
		\draw [->, color=blue] (y-3) -- (x-1);
		\draw [->, color=red] (y-1) -- (x-2);
		\draw [->, color=blue] (y-3) -- (x-3);
		\draw [->, color=red] (y-1) -- (x-4);
		\draw [->, color=green] (y-2) -- (x-5);
		\end{tikzpicture}
		\caption{}\label{fig:fmm:parent_sampling}
	\end{subfigure}
	\hspace{1cm}
	\begin{subfigure}[t]{0.2\textwidth}
		\centering
		\begin{tikzpicture}[every node/.style={draw,circle}]
		\foreach \x/\p in {1/4, 2/1, 3/5, 4/2, 5/3} {
			\node (x-\x) at (\p * 0.8,0) {$x_\x$};
		}
		\foreach \y in {1, 2, 3}
			\node (y-\y) at (\y+0.3333, 1.5) {$y_\y$};
		\draw [->] (y-3) -- (x-1);
		\draw [->] (y-1) -- (x-2);
		\draw [->] (y-3) -- (x-3);
		\draw [->] (y-1) -- (x-4);
		\draw [->] (y-2) -- (x-5);
		\end{tikzpicture}
		\caption{}\label{fig:fmm:sampled_forest}
	\end{subfigure}
	\caption{Visualization of sampling from a forest mixture model. (\subref{fig:fmm:random_edges}) In a forest mixture model, the edges between $\bm x$ and $\bm y$ are unknown random variables. (\subref{fig:fmm:parent_sampling}) To sample from the model, first the parent of each $x_i$ is chosen independently at random according to $p(\bm e_i)$. In this visualization, each $p(e_i)$ is uniform over the latent variables. (\subref{fig:fmm:sampled_forest}) After sampling a forest structure from $p(\bm e)$, $\bm x$ and $\bm y$ are sampled according to the resulting forest model.}
	\label{fig:fmm}
	\end{figure*}
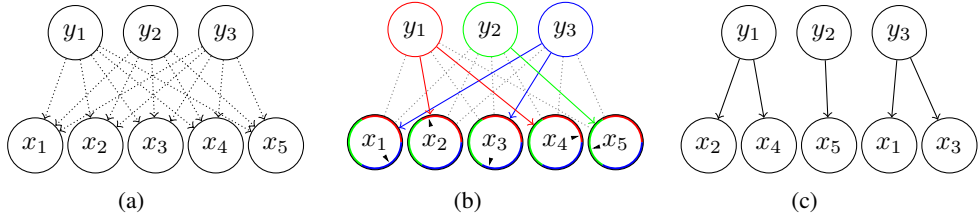
Consider a general directed model with a single layer of observed variables $\bm x \in \mathbb R^n$ and latent variables $\bm y \in \mathbb R^m$. The joint distribution $p(\bm x, \bm y)$ takes the form

\[
p(\bm x, \bm y) = \left[ \prod_{j=1}^m p(y_j) \right] \left[ \prod_{i=1}^n p(x_i | \bm y) \right]
\] 

A directed model is a \textit{forest model} if each $x_i$ has exactly one parent in the model's directed dependency graph; they are so-named because the resulting graphical model is a forest with one tree per latent variable $y_j$. These models are particularly simple because the $y_j$'s are conditionally independent given $\bm x$. Let $\bm e_i \in I^m$ be the one-hot vector indicating the parent of $x_i$, so $e_{ij} = 1$ if and only if $y_j$ is the parent of $x_i$. Then we can write

\[
p(x_i | \bm y) = \prod_{j=1}^m p(x_i|y_j)^{e_{ij}}
\]

Suppose we want to fit a forest model to data, but we don't know which $x_i$'s should be the children of which $y_j$'s. One way to handle this uncertainty is to treat the $\bm e_i$'s as independent latent random variable that have to be inferred, just like $\bm y$. To do this, we must first define a prior $p(\bm e_i)$ for each $i$. Given such a prior, the joint distribution over $\bm x$, $\bm y$, and $\bm e \equiv \{\bm e_i\}_{i=1}^n$ is

\[
p(\bm x, \bm y, \bm e) = \left[ \prod_{i=1}^n p(\bm e_i) \right] \left[ \prod_{j=1}^m p(y_j) \right] \left[ \prod_{i=1}^n p(x_i | \bm y, \bm e_i) \right]
\]

The resulting model is a \textit{forest mixture model} (FMM): to sample from this model, we first draw a random forest structure by sampling from the prior $p(\bm e)$; then, $\bm x$ and $\bm y$ are sampled from the selected forest model.

Though the $y_j$'s are no longer conditionally independent given $\bm x$, they are independent given $\bm x$ and $\bm e$. Similarly, the $\bm e_i$'s are conditionally independent given $\bm x$ and $\bm y$. To see this, define $\hat p(x_i | y_j) \equiv p(x_i | y_j, e_{ij}=1)$. Then the joint distribution can be written

\[
p(\bm x, \bm y, \bm e) = \left[ \prod_{i=1}^n p(\bm e_i) \right] \left[ \prod_{j=1}^m p(y_j) \right] \prod_{i=1}^n \prod_{j=1}^m \hat p(x_i | y_j)^{e_{ij}}
\]

In the next section, we will use the mean-field variational ELBO for this model, which for a given variational distribution $q(\bm y, \bm e | \bm x)$ is

\begin{align}
\log p(\bm x) &\geq \mathbb E [\log p(\bm x | \bm y, \bm e)] - D_{KL}(q(\bm y, \bm e | \bm x) \| p(\bm y, \bm e)) \nonumber \\
&= \sum_{i=1}^n \sum_{j=1}^m \mathbb E [e_{ij}] \mathbb E [\log \hat p(x_i | y_j)]  \nonumber \\
&- \sum_{j=1}^m D_{KL}(\left. q(y_j|\bm x) \right\Vert p(y_j)) \nonumber \\
&- \sum_{i=1}^n D_{KL}(\left. q(\bm e_i|\bm x) \right\Vert p(\bm e_i)) \label{eq:fmm_elbo}
\end{align}

\section{THE FOREST MIXTURE BOUND}
For simplicity, we only consider shallow models in this section. The extension to deep models is straightforward (see Appendix C).

A single-layer \textit{deep exponential family} (DEF) model is a directed model with a single layer of observed variables $\bm x \in \mathbb R^n$ and hidden variables $\bm y \in \mathbb R^m$, where the conditional distribution is in an exponential family. The joint distribution $p(\bm x, \bm y)$ takes the form

\begin{equation*} \begin{split}
p(\bm x, \bm y) &= \left[ \prod_{j=1}^m p(y_j) \right] \left[ \prod_{i=1}^n p(x_i | \bm y) \right] \\
p(x_i | \bm y) &= \exp \left\{ g(x_i) + t(x_i) \eta_i(\bm y) - a(\eta_i(\bm y)) \right\}
\end{split} \end{equation*}

Suppose we are given an observation $\bm x$ and want to approximately infer the posterior $p(\bm y | \bm x)$ by maximizing the variational ELBO, and suppose the $y_j$'s are conditionally independent given $\bm x$, so $p(\bm x, \bm y) = p(\bm x) \prod_{j=1}^m p(y_j | \bm x)$. Then the mean-field variational ELBO is

\begin{equation*} \begin{split}
\log p(x) &\geq \max_{q(\bm y | \bm x)} \mathbb E[\log p(\bm x, \bm y)] + H(\bm y | \bm x) \\
&\equiv \max_{q(\bm y | \bm x)} \sum_{j=1}^m \mathbb E \left[ \log p(y_j | \bm x) \right] + H(y_j | \bm x) \\
&= \sum_{j=1}^m  \max_{q(y_j | \bm x)}  \mathbb E \left[ \log p(y_j | \bm x) \right] + H(y_j | \bm x) \\ 
\end{split} \end{equation*}

In the second line, $\log p(\bm x)$ is constant with respect to $q(\bm y | \bm x)$ and can be removed without changing the optimization problem. In this case, the ELBO separates into a sum of terms, each of which involves only a single $y_j$. This allows us to optimize the ELBO by updating each $q(y_j | \bm x)$ independently and in parallel.

In a general DEF, the $y_j$'s are not conditionally independent and the objective does not separate. However, without much manipulation, much of the ELBO does separate: for a single-layer DEF, the ELBO can be written

\begin{equation*} \begin{split}
& \log p(\bm x) \geq \mathbb E[\log p(\bm x, \bm y)] + H(\bm y | \bm x) \\
&= \sum_{i=1}^n \mathbb E[\log p(x_i | \bm y)] + \sum_{j=1}^m \mathbb E[\log p(y_j)] + H(y_j | \bm x) \\
\end{split} \end{equation*} 

So only the $\mathbb E[\log p(x_i | \bm y)]$ terms aren't separable. However, if $\eta_i$ is an affine function of $\bm y$, so $\eta_i \equiv b_i + \bm w_i \cdot \bm y$ for some $b_i \in \mathbb R$ and $\bm w_i \in \mathbb R^m$, then each $\mathbb E[\log p(x_i | \bm y)]$ term can be expanded

\begin{equation*} \begin{split}
& \mathbb E[\log p(x_i | \bm y)] = g(x_i) + t(x_i) \mathbb E[\eta_i] - \mathbb E[a(\eta_i)] \\
&= g(x_i) + t(x_i) \left( b_i + \bm w_i \cdot \mathbb E[\bm y] \right) - \mathbb E[a(b_i + \bm w_i \cdot \bm y)]
\end{split} \end{equation*}

From this we can see the only term left preventing the entire ELBO from separating is $\mathbb E_{q(\bm y | \bm x)}[-a(\eta_i(\bm y))]$, a high-dimensional expectation of the non-linear log-partition function. The one thing we know about the log-partition function in exponential families is that it's convex. This suggests we use Jensen's inequality to bound $\mathbb E[-a(\eta_i)]$. Note that using Jensen's to bring the expectation over $q$ inside $a$ gives an inequality in the wrong direction because $-a(\eta_i)$ is concave; to get a lower bound, we need to pull an expectation out from the inside of $a$. The derivation of the ELBO gives a hint on how to do this: recall

\begin{equation*} \begin{split}
\log p(x) &= \log \int p(x,y) dy \\
&= \log \int \frac{q(y|x)}{q(y|x)} p(x,y) dy \\
&= \log \mathbb E_{q(y|x)} \left[ \frac{p(x,y)}{q(y|x)} \right] \\
&\geq \mathbb E_{q(y|x)} \left[ \log \frac{p(x,y)}{q(y|x)} \right]
\end{split} \end{equation*}

In the same way, we will introduce a variational or auxiliary distribution inside the concave function $-a(\eta)$, then use Jensen's to pull it out. For each $i$, introduce an auxiliary discrete distribution over $m$ categories $\bm \varepsilon_i \in \Delta^{m-1}$, so

\begin{equation*}
\sum_{j=1}^m \varepsilon_{ij} = 1 \hspace{1cm} \varepsilon_{ij} \geq 0 \ \forall j \in [m]
\end{equation*}

Injecting this inside the log-partition function gives

\begin{equation*}
\mathbb E[-a(b_i + \bm w_i \cdot \bm y)] = \mathbb E\left[ -a \left( b_i + \sum_{j=1}^m \varepsilon_{ij} \frac{w_{ij} y_j}{\varepsilon_{ij}} \right) \right] \\
\end{equation*}

To use Jensen's inequality, we first need to bring $b_i$ inside the sum, which we can do using $b_i = \sum_{j=1}^m \varepsilon_{ij} b_i$. This partitions the bias $b_i$ into $m$ parts according to $\bm \varepsilon_i$. However, to get a sufficiently tight bound, we'll need to consider more general splittings: introduce another set of auxiliary parameters $\hat{\bm b}_i \in \mathbb R^m$ with the constraint $b_i = \sum_{j=1}^m \varepsilon_{ij} \hat b_{ij}$. Then

\begin{gather}
\mathbb E [-a(b_i + \bm w_i \cdot \bm y)] = \mathbb E \left[ -a \left( \sum_{j=1}^m \varepsilon_{ij} \left( \hat b_{ij} + \frac{w_{ij} y_j}{\varepsilon_{ij}} \right) \right) \right] \nonumber \\
\geq \sum_{j=1}^m \varepsilon_{ij} \mathbb E \left[ -a \left( \hat b_{ij} + \frac{w_{ij} y_j}{\varepsilon_{ij}} \right) \right] \label{eq:L_verbose}
\end{gather}

Bounding this term for each $i$ separates the entire ELBO into a sum of terms, each of which involves only a single $y_j$. Plugging this in directly to get a final bound on log-likelihood results in an unwieldy expression, so first we will introduce new notation to simplify the bound.

\subsection{CONNECTION WITH FMM}
To demonstrate the relation of the above bound and forest mixture models, let us define
\begin{equation*} \begin{split}
\hat \eta_{ij} &\equiv \hat b_{ij} + \frac{w_{ij} y_j}{\varepsilon_{ij}} \\
\hat p(x_i | y_j) &\equiv \exp \left\{ g(x_i) + t(x_i) \hat \eta_{ij} - a(\hat \eta_{ij}) \right\}
\end{split} \end{equation*}
Then $\eta_i = \sum_{j=1}^m \varepsilon_{ij} \hat \eta_{ij}$ and the bound can be rewritten as follows:
\begin{equation}
\mathbb E \left[ -a \left( \eta_i \right) \right] \geq \sum_{j=1}^m \varepsilon_{ij} \mathbb E \left[ -a \left( \hat \eta_{ij} \right) \right] \label{eq:L_eta}
\end{equation}
This expression can be used to impose bounds on each $\mathbb E[\log p(x_i | \bm y)]$:
\begin{equation*} \begin{split}
&\mathbb E [\log p(x_i | \bm y)] = g(x_i) + t(x_i) \mathbb E [\eta_i] - \mathbb E [a(\eta_i)] \\
&\geq g(x_i) + t(x_i) \mathbb E [\eta_i] - \sum_{j=1}^m \varepsilon_{ij} \mathbb E[a(\hat \eta_{ij})] \\
&= \sum_{j=1}^m \varepsilon_{ij} \left( g(x_i) + t(x_i) \mathbb E [\hat \eta_{ij}] -  \mathbb E[a(\hat \eta_{ij})] \right) \\
&= \sum_{j=1}^m \varepsilon_{ij} \mathbb E[\log \hat p(x_i|y_j)]
\end{split} \end{equation*}
Finally, plugging the above expression into the ELBO gives
\begin{align}
\log p(x) &\geq \mathbb E[\log p(\bm x, \bm y)] + H (\bm y | \bm x) \nonumber \\
&\geq \sum_{i=1}^n \sum_{j=1}^m \varepsilon_{ij} \mathbb E[\log \hat p(x_i | y_j)] \nonumber \\
&- \sum_{j=1}^m D_{KL}(q(\bm y | \bm x) \| p(\bm y)) \label{eq:fm_bound}
\end{align}
Comparing \eqref{eq:fm_bound} with \eqref{eq:fmm_elbo} confirms that this bound is {\em identical} to the ELBO of a forest mixture model with the same $\hat p(x_i, y_j)$ and $q(y_j | \bm x)$, with $q(e_{ij} = 1 | \bm x) = \varepsilon_{ij}$ (so that $\mathbb E[e_{ij}] = \varepsilon_{ij}$) and $p(\bm e_i) = q(\bm e_i | \bm x)$ (so that the second $KL$ term of the FMM ELBO is zero and disappears entirely). For this reason, we call this bound the \textit{forest mixture bound} (FM bound). Note this bounds the DEF ELBO by the ELBO of each FMM in a large family of FMM's parameterized by $\bm \varepsilon \equiv \{\bm \varepsilon_i\}_{i=1}^n $ and $\hat{\bm b} \equiv \{\hat{\bm b}_i\}_{i=1}^n$.

\section{ALGORITHM}
\begin{algorithm}[t]
    \SetKwInOut{Input}{input}
    \SetKwInOut{Output}{output}
	\Input{An observation $\bm x \in \mathbb R^n$ and model parameters $W \in \mathbb R^{n \times m}$, $\bm b \in \mathbb R^n$, $\sigma_y^2 \in \mathbb R$ and $\sigma_x^2 \in \mathbb R$.}
	\Output{The mean-field variational distribution $q(\bm y | \bm x) \equiv \prod_{j=1}^m q(y_j | \bm x)$}
	initialize $(\mu_0)_j$ and $(\sigma_0)_j^2$ for each $j \in [m]$ \\
	\For{$t = 0$ \KwTo $T-1$}{
		\For{$i=1$ \KwTo $n$} {
			\For{$j=1$ \KwTo $m$} {
				$(\varepsilon_t)_{ij} = \frac{|w_{ij}| (\sigma_t)_j}{\sum_{j'=1}^m |w_{ij'}| (\sigma_t)_{j'}^2}$
				$(\hat{b}_t)_{ij} = (b_i + \sum_{j=1}^m w_{ij} (\mu_t)_j) - \frac{w_{ij} (\mu_t)_j}{ (\varepsilon_t)_{ij}}$
			}
		}
		\For{$j=1$ \KwTo $m$}{
			$(\mu_{t+1})_j \equiv \frac{\sum_{i=1}^m w_{ij} (x_i - (\hat{b}_t)_{ij})}{\frac{\sigma_x^2}{\sigma_y^2} + \sum_{j=1}^m \frac{w_{ij}^2}{(\varepsilon_t)_{ij}}}$
			$(\sigma_{t+1})_j^2 \equiv \frac{1}{\frac{1}{\sigma_y^2} + \frac{1}{\sigma_x^2} \sum_{j=1}^m \frac{w_{ij}^2}{(\varepsilon_t)_{ij}}}$
		}
	}
	\Return $q(y_j | \bm x) = \mathcal N((\mu_T)_j, (\sigma_T)_j^2)$ for $j \in [m]$
	\caption{The FM algorithm in the Gaussian case.}
	\label{alg:fm_algorithm}
\end{algorithm}
To optimize the FM bound, we propose an alternating maximization algorithm: in the first step, update all $q(y_j | \bm x)$ in parallel while holding all $\varepsilon_{ij}$ and $\hat b_{ij}$ fixed; in the second step, update all $\varepsilon_{ij}$ and $\hat b_{ij}$ in parallel while holding all $q(y_j | \bm x)$ fixed. In this section, we will derive the optimal updates for $p(y_j | \bm x)$, $\varepsilon_{ij}$, and $\hat b_{ij}$ in the case when each $x_i$ and $y_j$ are Gaussian with known variance:

\begin{equation*} \begin{split}
p(y_j) = \mathcal N(0, \sigma_y^2) \hspace{1cm} p(x_i|\bm y) = \mathcal N(\eta_i(\bm y), \sigma_x^2)
\end{split} \end{equation*}

We will derive the updates for the auxiliary parameters first since this will help simplify the update for the variational distribution later.
\subsection{AUXILIARY PARAMETER UPDATES}
Maximizing the FM bound over $\bm \varepsilon$ and $\hat{\bm b}$ is equivalent to maximizing $\mathcal L_i \equiv \sum_{j=1}^m \varepsilon_{ij} \mathbb E[-a(\hat \eta_{ij})]$ over $\bm \varepsilon_i$ and $\hat{\bm b}_i$ for each $i$, since these are the only terms in the FM bound that depend on $\bm \varepsilon$ and $\hat{\bm b}$. In the Gaussian case, $-a(\hat \eta_{ij}) = -\frac{1}{2 \sigma_x^2} \hat \eta_{ij}^2$ and

\begin{equation*} \begin{split}
& \mathcal L_i = \sum_{j=1}^m \varepsilon_{ij} \mathbb E \left[ -\frac{1}{2 \sigma_x^2} \hat \eta_{ij} ^2 \right] \\
&= -\frac{1}{2 \sigma_x^2} \sum_{j=1}^m \varepsilon_{ij} \left( \text{Var}\left[ \hat \eta_{ij} \right] + \mathbb E \left[ \hat \eta_{ij} \right]^2 \right) \\
&= -\frac{1}{2 \sigma_x^2} \sum_{j=1}^m \frac{w_{ij}^2 \text{Var}[y_j]}{\varepsilon_{ij}} + \varepsilon_{ij} \left( \hat b_{ij} + \frac{w_{ij} \mathbb E[y_j]}{\varepsilon_{ij}} \right)^2
\end{split} \end{equation*}

\begin{theorem}
Holding $q(y_j | \bm x)$ constant, the choice of $\hat{\bm b}_i$ and $\bm \varepsilon_i$ that maximizes $\mathcal L_i$ is $\hat{\bm b}_i = \hat{\bm b}_i^*$ and $\bm \varepsilon_i = \bm \varepsilon_i^*$, where
\begin{align*}
\hat b_{ij}^* &= \mathbb E[\eta_i] - \frac{w_{ij} \mathbb E[y_j]}{\varepsilon_{ij}^*} & 
\varepsilon_{ij}^* &= \frac{|w_{ij}| \text{Std}[y_j]}{\sum_{j'=1}^m |w_{ij'}| \text{Std}[y_{j'}]}
\end{align*}
\end{theorem}

For a proof, see Appendix A. Note that these computations can be parallelized across $i$ and $j$.

\subsection{VARIATIONAL UPDATES}
Holding the auxiliary parameters fixed, each variational distribution $q(y_j | \bm x)$ can be updated in parallel:

\begin{theorem}
For a fixed $\bm \varepsilon$ and $\hat{\bm b}$, the choice for the next variational distribution $q_{t+1}(y_j | \bm x)$ that maximizes the FM bound is $q_{t+1}(y_j | \bm x) = \mathcal N((\mu_{t+1}^*)_j, (\sigma_{t+1}^*)_j^2)$, where

\begin{gather*}
(\mu_{t+1}^*)_j \equiv \frac{(\bm x - \mathbb E_{q_t}[\bm \eta]) \cdot \bm w_j + \mathbb E_{q_t}[y_j] \sum_{i=1}^n \frac{w_{ij}^2}{\varepsilon_{ij}}}{\frac{\sigma_x^2}{\sigma_y^2} + \sum_{i=1}^n \frac{w_{ij}^2}{\varepsilon_{ij}}} \\
(\sigma_{t+1}^*)_j^2 \equiv \frac{1}{\frac{1}{\sigma_y^2} + \frac{1}{\sigma_x^2} \sum_{i=1}^n \frac{w_{ij}^2}{\varepsilon_{ij}}}
\end{gather*}
\end{theorem}

For a proof, see Appendix B.

\section{DISCUSSION}
\paragraph{Tightness}

We derived the FM bound by using Jensen's inequality to lower bound the ELBO. For a given variational distribution $q$, the gap between the two bounds is

\[
\text{GAP} \equiv \sum_{i=1}^n \mathbb E[-a(\eta_i)] - \sum_{i=1}^n \sum_{j=1}^m \varepsilon_{ij} \mathbb E[-a(\hat \eta_{ij})]
\] 

In the Gaussian case, for an optimal choice of auxiliary parameters (see Appendix A),

\begin{gather*}
\sum_{j=1}^m \varepsilon_{ij} \mathbb E[-a(\hat \eta_{ij})] = -\frac{1}{2 \sigma_x^2} \| \bm w_i \odot \text{Std}[\bm y]\|_1^2 - \frac{1}{2 \sigma_x^2} \mathbb E[\eta_i]^2 \\
\mathbb E[-a(\eta_i)] = -\frac{1}{2 \sigma_x^2} \| \bm w_i \odot \text{Std}[\bm y]\|_2^2 - \frac{1}{2 \sigma_x^2} \mathbb E[\eta_i]^2 \\
\text{GAP} = \frac{1}{2 \sigma_x^2} \sum_{i=1}^n \| \bm w_i \odot \text{Std}[\bm y]\|_1^2 - \| \bm w_i \odot \text{Std}[\bm y]\|_2^2
\end{gather*}

Since $\sum_{i=1}^n \|\bm w_i\|_1^2 \geq \sum_{i=1}^n \|\bm w_i\|_2^2$, the FM bound imposes a stronger regularization on the variance of the variational distribution compared to the variational ELBO. For this reason, the variational distribution $q$ that maximizes the FM bound generally has a smaller variance compared to the variational distribution that maximizes the ELBO.


The FM bound tightly bounds the ELBO when $p$ is a forest model, so that $w_{ij}$ has exactly one non-zero element in the component $j(i)$ corresponding to the parent of $x_i$. In this case, 

\begin{align*}
\|\bm w_i \odot \text{Std}[\bm y]\|_1^2 = w_{ij(i)}^2 \text{Var}[y_{j(i)}] = \|\bm w_i \odot \text{Std}[\bm y]\|_2^2
\end{align*}

The bound is also tight when $\text{Var}[\bm y] = 0$, but in this case both the ELBO and the FM bound yield $-\infty$ because of the conditional entropy term $H(\bm y | \bm x)$.

\paragraph{Speed of Convergence}

Let's examine the role of $\bm \varepsilon$ in the update for $q(y_j | \bm x)$. If $\sum_{j=1}^m \frac{w_{ij}^2}{\varepsilon_{ij}}$ is large, then $\mathbb E_{q_{t+1}} \approx \mathbb E_{q_t}[y_j]$, and so the FM algorithm makes a small update for $y_j$. If $\sum_{j=1}^m \frac{w_{ij}^2}{\varepsilon_{ij}}$ is small, then $\mathbb E_{q_{t+1}}[y_j]$ makes a large step in the direction of the residual \mbox{$\bm x - \mathbb E[\bm \eta]$}. In fact, if for some $j$, $\varepsilon_{ij} = 1$ for all $i$ where $w_{ij}$ is non-zero, then the FM algorithm updates $q(y_j | \bm x)$ exactly as CAVI would. In this sense, $\bm \varepsilon$ acts like an attention parameter that selects which $q(y_j | \bm x)$ to change and by how much.

If $p$ is a forest model, then the FM algorithm chooses $\bm \varepsilon_i$ to be the one-hot vector indicating the parent of $x_i$. In this case, the FM algorithm makes coordinate updates for all $j$ in parallel and converges in one iteration. If $p$ is forest-like, i.e., $|\bm w_j| \cdot |\bm w_{j'}|$ is small for $j \neq j'$, then $\bm \varepsilon_i$ is close to one-hot and the FM algorithm makes damped, nearly-CAVI updates in parallel. In this sense, the speed at which the FM algorithm converges depends on how inherently forest-like the model $p$ is.

\section{EXPERIMENTS}
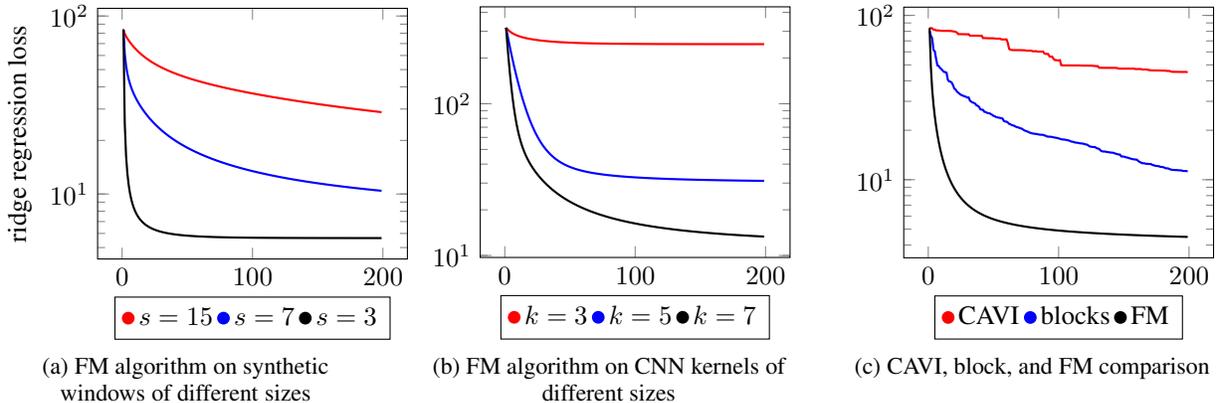
\begin{figure*}[t]
	\captionsetup[subfigure]{justification=centering}
	\centering
	\begin{subfigure}[t]{0.3\textwidth}
		\centering
		\begin{tikzpicture}
		\begin{semilogyaxis}[
			xlabel={iterations},
			ylabel={ridge regression loss},
			cycle list name=color list,
			legend pos=north east,
			every axis plot/.append style={thick},
			width=5.7cm,
			legend style={at={(0.5,-0.15)},
			anchor=north,legend columns=-1}
		]
		\addlegendimage{only marks, red, mark=*}
		\addlegendimage{only marks, blue, mark=*}
		\addlegendimage{only marks, black, mark=*}
		\addplot [red] table [x expr=\thisrowno{0}, y expr=\thisrowno{1}*-1] {data/fm_windows15.dat};
		\addplot [blue] table [x expr=\thisrowno{0}, y expr=\thisrowno{1}*-1] {data/fm_windows7.dat};
		\addplot [black] table [x expr=\thisrowno{0}, y expr=\thisrowno{1}*-1] {data/fm_windows3.dat};
		\legend{$s=15$,$s=7$,$s=3$}
		\end{semilogyaxis}
		\end{tikzpicture}
		\subcaption{FM algorithm on synthetic windows of different sizes}	
	\end{subfigure}
	\hspace{0.5cm}
	\begin{subfigure}[t]{0.3\textwidth}
		\centering
		\begin{tikzpicture}
		\begin{semilogyaxis}[
			xlabel={iterations},
			cycle list name=color list,
			legend pos=north east,
			every axis plot/.append style={thick},
			width=5.7cm,
			legend style={at={(0.5,-0.15)},
			anchor=north,legend columns=-1}
		]
		\addlegendimage{only marks, red, mark=*}
		\addlegendimage{only marks, blue, mark=*}
		\addlegendimage{only marks, black, mark=*}
		\addplot [red] table [x expr=\thisrowno{0}, y expr=\thisrowno{1}*-1] {data/fm_conv3.dat};
		\addplot [blue] table [x expr=\thisrowno{0}, y expr=\thisrowno{1}*-1] {data/fm_conv5.dat};
		\addplot [black] table [x expr=\thisrowno{0}, y expr=\thisrowno{1}*-1] {data/fm_conv7.dat};
		\legend{$k=3$,$k=5$,$k=7$}
		\end{semilogyaxis}
		\end{tikzpicture}
		\subcaption{FM algorithm on CNN kernels of different sizes}	
	\end{subfigure}
    \hspace{0.5cm}
    \begin{subfigure}[t]{0.3\textwidth}
		\centering
		\begin{tikzpicture}
		\begin{semilogyaxis}[
			xlabel={iterations},
			cycle list name=color list,
			legend pos=north east,
			every axis plot/.append style={thick},
			width=5.7cm,
			legend style={at={(0.5,-0.15)},
			anchor=north,legend columns=-1}
		]
		\addlegendimage{only marks, red, mark=*}
		\addlegendimage{only marks, blue, mark=*}
		\addlegendimage{only marks, black, mark=*}
		\addplot [red] table [x expr=\thisrowno{0}, y expr=\thisrowno{1}*-1] {data/blocks1.dat};
		\addplot [blue] table [x expr=\thisrowno{0}, y expr=\thisrowno{1}*-1] {data/blocks16.dat};
		\addplot [black] table [x expr=\thisrowno{0}, y expr=\thisrowno{1}*-1] {data/block_free.dat};
		\legend{CAVI,blocks,FM}
		\end{semilogyaxis}
		\end{tikzpicture}
		\subcaption{CAVI, block, and FM comparison}	
	\end{subfigure}
    \caption{The ridge regression objective over 200 iterations.}
\end{figure*}
Recall that we derived the FM bound by lower bounding the ELBO. Algorithms that optimize the ELBO like CAVI will generally provide a superior lower bound on log-likelihood compared to the FM algorithm. For a more fair comparison, we can instead measure how quickly these algorithms converge to the optimal mean. In the Gaussian case, optimizing the mean of the mean-field variational distribution is equivalent to minimizing a ridge regression objective:

\[
\frac{1}{2 \sigma_x^2} \sum_{i=1}^n \left( x_i - (b_i + \bm w_i \cdot \mathbb E[\bm y]) \right)^2 + \frac{1}{2 \sigma_y^2} \sum_{j=1}^m \mathbb E[y_j]^2
\]

To evaluate each algorithm on the ridge regression problem, we must first choose a $\bm x$, $\bm b$, and a set of $w_{ij}$. All the algorithms we consider in this section are guaranteed to converge to the optimal solution, so we are only interested in comparing how  quickly each algorithm converges to that optimal solution. This is measured by recording the objective value achieved by the mean of the variational distribution $\mathbb E_{q_t}[\bm y]$ in the ridge regression problem across $200$ iterations.

In the first experiment, we choose $\bm x$ to be a vectorized sample from the MNIST dataset, with pixel values scaled to lie in the interval $[-1, 1]$; we choose $\bm b$ to be the average of $1000$ randomly chosen MNIST samples; and we construct a synthetic $w_{ij}$ as follows: given an integer window side length $s$, we construct all possible square $s \times s$ windows of pixels. For windows that overlap the border of the $28 \times 28$ MNIST image region, we clip the window so that it lies entirely inside the image region, resulting in a rectangular window. For each window, we add a latent variable $y_j$ to the model and a corresponding $\bm w_j$, where $w_{ij} = 1$ if pixel $i$ lies in window $j$, and $w_{ij} = 0$ otherwise. The resulting model is more forest-like for smaller choices of $s$: if $s = 1$, the windows are disjoint and the graphical model is exactly a forest. Figure 2a demonstrates the rate of convergence of the FM algorithm for various choices of $s$. As we expect, the FM algorithm converges faster for more forest-like models, i.e., smaller $s$. Note that the objective value achieved by the optimal solution to the ridge regression problem changes as $w_{ij}$ changes.

The second experiment is similar to the first, except it uses $\bm x$ from the CIFAR-10 dataset, $\bm b = 0$, and instead of uniform windows, uses the first layer kernels from a convolutional neural net trained several times changing only the width of the first layer kernels. Figure 2b demonstrates the FM algorithm converges faster for more forest-like models even using real-world data.

Our last experiment compares the convergence of the FM algorithm with CAVI and block coordinate ascent. Here we choose $\bm x$ and $\bm b$ the same as in the first experiment, but we choose $w_{ij}$ differently to make blocking the latent variables easy: first we partition the $28 \times 28$ MNIST image region into $16$ regions, each of size $7 \times 7$. Then, we construct all possible $7 \times 7$ windows (as in the first experiment with $s=7$), then clip them to fit in the first region. This is repeated for each region. If we block the latent variables according to which region the corresponding windows were clipped to, then the blocks will be conditionally independent, since windows clipped to different regions must be disjoint. Blocking in this way guarantees that the block coordinate algorithm will converge to the optimal solution. Figure 2c compares the rate of convergence for CAVI, block coordinate ascent, and the FM algorithm. The figure shows our block-free method can outperform the block coordinate method, even when the blocking is quite good.

\section{CONCLUSION}
In this paper we derived a forest mixture bound on the log-likelihood of deep exponential families. This bound gets around the ``explaining away effect'' by using a set of auxiliary parameters to separate the problem of inference for each latent variable in the same layer, allowing us to make parallel updates. We then made a deep dive into the simple case where all variables are Gaussian: we derived the exact variable updates, then tested the algorithm on both synthetic and real-world data. Our promising results show that fast, parallel inference in deep exponential families is possible without the use of blocks.

\appendix
\section{AUXILIARY PARAMETER UPDATES}
Proof of {\em \bf{Theorem 1}}: First, we will find the optimal choice of $\hat{\bm b}_i$ for any given $\bm \varepsilon_i$. Since $\hat{ \bm b}_i$ is constrained by $\sum_{j=1}^m \varepsilon_{ij} \hat b_{ij} = b_i$, let's first parameterize $\hat{\bm b}_i$ by a set of unconstrained parameters: let $\bm \gamma_{i} \in \mathbb R^m$ and write

\[
\hat b_{ij} = b_i - \gamma_{ij} + \bm \varepsilon_i \cdot \bm \gamma_i
\]

So for any choice of $\bm \gamma_i$, the constraint $b_i = \sum_{j=1}^m \varepsilon_{ij} \hat b_{ij}$ is satisfied. Now we can differentiate the bound with respect to $\gamma_{ij}$, set to zero and solve. We will need the following partial derivatives:

\begin{align*}
\frac{\partial \hat b_{ij}}{\partial \gamma_{ij}} &= -1 + \varepsilon_{ij} & \frac{\partial \hat b_{ij'}}{\partial \gamma_{ij}} &= \varepsilon_{ij} \ \forall j' \neq j
\end{align*}

Now setting the partial derivative of $\mathcal L_i$ with respect to $\gamma_{ij}$ to zero,

\begin{align*}
0 = \frac{\partial}{\partial \gamma_{ij}} \mathcal L_i &= - \frac{1}{\sigma_x^2} \sum_{j'=1}^m \varepsilon_{ij'} \mathbb E[\hat \eta_{ij'}] \frac{\partial \hat b_{ij'}}{\partial \gamma_{ij}} \\
&= \frac{\varepsilon_{ij}}{\sigma_x^2} \left( \mathbb E[\hat \eta_{ij}] - \sum_{j'=1}^m \varepsilon_{ij'} \mathbb E[\hat \eta_{ij'}] \right)
\end{align*}

The derivative is zero for all $j$ in particular when the choice of $\hat b_{ij}$ makes $\mathbb E[\hat \eta_{ij}]$ constant across $j$. We can verify this is satisfied by the choice $\gamma_{ij} = \frac{w_{ij}}{\varepsilon_{ij}} \mathbb E[y_j]$, which makes $\hat b_{ij} = \hat b_{ij}^*$:

\begin{align*}
\mathbb E[\hat \eta_{ij}] &= \mathbb E \left[ \hat b_{ij} + \frac{w_{ij}}{\varepsilon_{ij}} y_j \right] \\
&= \mathbb E \left[ \mathbb E[\eta_i] - \frac{w_{ij}}{\varepsilon_{ij}} \mathbb E[y_j] + \frac{w_{ij}}{\varepsilon_{ij}} y_j \right] \\
&= \mathbb E[\eta_i]
\end{align*}

Plugging this choice into $\mathcal L_i$ yields

\begin{align*}
\mathcal L_i &= -\frac{1}{2 \sigma_x^2} \sum_{j=1}^m \left( \frac{w_{ij}^2 \text{Var}[y_j]}{\varepsilon_{ij}} + \varepsilon_{ij} \mathbb E[\eta_i]^2 \right) \\
&= - \frac{1}{2 \sigma_x^2} \left( \sum_{j=1}^m \frac{w_{ij}^2 \text{Var}[y_j]}{\varepsilon_{ij}} \right) - \frac{1}{2 \sigma_x^2} \mathbb E[\eta_i]^2
\end{align*}

Now let's try to find the optimal choice of $\varepsilon_{ij}$. Since $\varepsilon_{ij}$ is constrained by $\varepsilon_i \in \Delta^{m-1}$, we'll also parameterize $\varepsilon_{ij}$ by a set of unconstrained parameters $\bm \tau_i \in \mathbb R^m$: 

\[
\varepsilon_{ij} = \exp\{\tau_{ij}\} / \sum_{j'=1}^m \exp \{ \tau_{ij'} \}
\]

We will need the following partial derivatives:

\begin{align*}
\frac{\partial \varepsilon_{ij}}{\partial \tau_{ij}} &= \varepsilon_{ij}(1 - \varepsilon_{ij}) & \frac{\partial \varepsilon_{ij'}}{\partial \tau_{ij}} &= -\varepsilon_{ij} \varepsilon_{ij'} \  \forall j' \neq j
\end{align*}

Now setting the partial derivative of $\mathcal L_i$ with respect to $\tau_{ij}$ to zero,

\begin{align*}
0 = \frac{\partial}{\partial \tau_{ij}} \mathcal L_i &= \frac{1}{2 \sigma_x^2} \sum_{j'=1}^m \frac{w_{ij'}^2 \text{Var}[y_{j'}]}{\varepsilon_{ij'}^2} \frac{\partial \varepsilon_{ij'}}{\partial \tau_{ij}} \\
&= \frac{1}{2 \sigma_x^2} \sum_{j'=1}^m \text{Var}[\hat \eta_{ij'}] \frac{\partial \varepsilon_{ij'}}{\partial \tau_{ij}} \\
&= \frac{\varepsilon_{ij}}{2 \sigma_x^2} \left( \text{Var}[\hat \eta_{ij}] - \sum_{j'=1}^m \varepsilon_{ij'} \text{Var}[\hat \eta_{ij'}] \right) \\
\end{align*}

The derivative is zero for all $j$ in particular when the choice of $\varepsilon_{ij}$ makes $\text{Var}[\hat \eta_{ij}]$ constant across $j$. We can verify this is satisfied by the choice $\tau_{ij} = \log |w_{ij} \text{Std}[y_j]|$, which makes $\varepsilon_{ij} = \varepsilon_{ij}^*$:

\begin{align*}
\text{Var}[\hat \eta_{ij}] &= \frac{w_{ij}^2 \text{Var}[y_j]}{\varepsilon_{ij}^2} \\
&= \frac{w_{ij}^2 \text{Var}[y_j]}{w_{ij}^2 \text{Var}[y_j] / \left(\sum_{j'=1}^m |w_{ij'}| \text{Std}[y_{j'}] \right)^2} \\
&= \| \bm w_i \odot \text{Std}[\bm y]\|_1^2
\end{align*}

Plugging this choice into $\mathcal L_i$ yields

\begin{align*}
\mathcal L_i &= -\frac{1}{2 \sigma_x^2} \sum_{j=1}^m |w_{ij}| \text{Std}[y_j] \left( \sum_{j'=1}^m |w_{ij}| \text{Std}[y_j] \right) \\
& \hspace{5cm} - \frac{1}{2 \sigma_x^2} \mathbb E[\eta_i]^2 \\
&= -\frac{1}{2 \sigma_x^2} \left( \sum_{j=1}^m |w_{ij}| \text{Std}[y_j] \right)^2 - \frac{1}{2 \sigma_x^2} \mathbb E[\eta_i]^2 \\
&= -\frac{1}{2 \sigma_x^2} \| \bm w_i \odot \text{Std}[\bm y] \|_1^2 - \frac{1}{2 \sigma_x^2} \mathbb E[\eta_i]^2
\end{align*} 
\section{VARIATIONAL UPDATES}
Proof of {\em \bf{Theorem 2}}: First, note that for any DEF, the optimal update equation is
as follows:
\begin{equation}
\log q_{t+1}(y_j | \bm x) \propto \log p(y_j) + \sum_{i=1}^n \varepsilon_{ij} \log \hat p_t(x_i|y_j)
\end{equation}
In the Gaussian case, we have
\begin{equation*} \begin{split}
\log p(y_j) &\propto -\frac{1}{2 \sigma_y^2} y_j^2 \\
\log \hat p(x_i | y_j) &\propto -\frac{1}{2 \sigma_x^2} (x_i - \hat \eta_{ij})^2 \\
&\propto -\frac{1}{2 \sigma_x^2} \left( x_i - \left( \hat b_{ij} + \frac{w_{ij}}{\varepsilon_{ij}} y_j \right) \right)^2 \\
&\propto \frac{1}{\sigma_x^2} \frac{(x_i - \hat b_{ij}) w_{ij}}{\varepsilon_{ij}} y_j - \frac{1}{2 \sigma_x^2} \frac{w_{ij}^2}{\varepsilon_{ij}^2} y_j^2 \\
\end{split} \end{equation*}

Plugging this in yields
\begin{gather*} \begin{split}
\log q(y_j | \bm x) &\propto \frac{1}{\sigma_x^2} \left( (\bm x - \hat{\bm b}_j) \cdot \bm w_j \right) y_j \\
&- \frac{1}{2} y_j^2 \left(\frac{1}{\sigma_y^2} + \frac{1}{\sigma_x^2} \sum_{i=1}^n \frac{w_{ij}^2}{\varepsilon_{ij}}\right)
\end{split} \\
\propto -\frac{1}{\sigma_x^2} \left( (\bm x - \hat{\bm b}_j) \cdot \bm w_j \right) y_j  - \frac{1}{2 (\sigma_{t+1}^*)_j^2} y_j^2 \\
\propto -\frac{1}{2 (\sigma_{t+1}^*)_j^2} \left( y_j - \frac{\frac{1}{\sigma_x^2} (\bm x - \hat{\bm b}_j) \cdot \bm w_j}{\frac{1}{\sigma_y^2} + \frac{1}{\sigma_x^2} \sum_{i=1}^n \frac{w_{ij}^2}{\varepsilon_{ij}}} \right)^2 \\
\propto -\frac{1}{2 (\sigma_{t+1}^*)_j^2} \left( y_j - \frac{(\bm x - \hat{\bm b}_j) \cdot \bm w_j}{\frac{\sigma_x^2}{\sigma_y^2} + \sum_{i=1}^n \frac{w_{ij}^2}{\varepsilon_{ij}}} \right)^2 \\
\end{gather*}

After substituting $\hat b_{ij} = \mathbb E[\eta_i] - \frac{w_{ij}}{\varepsilon_{ij}} \mathbb E_{q_t}[y_j]$ and rearranging, we get $\log q_{t+1}(y_j | \bm x) \propto \mathcal N((\mu_{t+1}^*)_j, (\sigma_{t+1}^*)_j^2)$.
\section{EXTENSION TO DEEP MODELS}
A DEF model with observed variables $\bm y^{(0)} \in \mathbb R^{m_0}$ and $L$ layers of latent variables $\{\bm y^{(\ell)}\}_{\ell=1}^L$ with $\bm y^{(\ell)} \in \mathbb R^{m_\ell}$ has joint distribution

\begin{align*}
p(\{\bm y^{(\ell)} \}_{\ell=0}^L) &= \left[ \prod_{\ell=0}^{L-1} \prod_{i=1}^{m_\ell} p(y^{(\ell)}_i | \bm y^{(\ell+1)}) \right] \left[ \prod_{i=1}^{m_L} p(y^{(L)}_i) \right] \\
p(y^{(\ell)}_i | \bm y^{(\ell+1)}) &= \exp \left\{ g(y^{(\ell)}_i) + t(y^{(\ell)}_i) \eta^{(\ell)}_i - a(\eta^{(\ell)}_i) \right\} \\
\eta^{(\ell)}_i &\equiv b^{(\ell)}_i + \bm w^{(\ell)}_i \cdot \bm y^{(\ell+1)}
\end{align*}

The ELBO for this model is

\begin{align*}
\log p(y^{(0)}) &\geq \sum_{i=1}^{m_0} \mathbb E [\log p(y^{(0)}_i | \bm y^{(1)})] \\
&+ \sum_{\ell=1}^{L-1} \sum_{i=1}^{m_\ell} \mathbb E [\log p(y^{(\ell)}_i | \bm y^{(\ell+1)})] + \underset{q}{H} (y^{(\ell)}_i | \bm y^{(0)}) \\
&+ \sum_{i=1}^{m_L} p(y^{(L)}_i) + \underset{q}{H} (y^{(L)}_i | \bm y^{(0)})
\end{align*}

For each $\ell \in \{0, \dots, L-1\}$, introduce the auxiliary parameters $\{ \bm \varepsilon^{(\ell)}_{i}\}_{i=1}^{m_\ell}$ and $\{ \hat {\bm b}^{(\ell)}_{i} \}_{i=1}^{m_\ell}$, with $\bm \varepsilon^{(\ell)}_i \in \Delta^{m_{\ell+1}-1}$ and $\hat {\bm b}^{(\ell)}_i \in \mathbb R^{m_{\ell+1}}$ constrained by $b^{(\ell)}_i = \sum_{j=1}^{m_{\ell+1}} \varepsilon^{(\ell)}_{ij} \hat b^{(\ell)}_{ij}$. For all $\ell \in \{0, \dots, L-1\}$, $i \in [m_\ell]$, and $j \in [m_{\ell+1}]$, define

\begin{align*}
\hat \eta^{(\ell)}_{ij} &\equiv \hat b^{(\ell)}_{ij} + \frac{w^{(\ell)}_{ij}}{\varepsilon^{(\ell)}_{ij}} y^{(\ell+1)}_j \\
\hat p(y^{(\ell)}_i | y^{(\ell+1)}_j) &\equiv \exp \{ g(y^{(\ell)}_i) + t(y^{(\ell)}_i) \hat \eta^{(\ell)}_{ij} - a(\hat \eta^{(\ell)}_{ij}) \}
\end{align*}

Then by (\ref{eq:L_eta}),

\begin{align*}
\mathbb E [\log p(y^{(\ell)}_i | \bm y^{(\ell+1)})] &\geq \sum_{j=1}^{m_{\ell+1}} \varepsilon^{(\ell)}_{ij} \mathbb E [\log \hat p(y^{(\ell)}_i | y^{(\ell+1)}_j)]
\end{align*}

Plugging this into the ELBO yields

\begin{align*}
& \log p(\bm y^{(0)}) \geq \sum_{i=1}^{m_0} \sum_{j=1}^{m_1} \varepsilon^{(\ell)}_{ij} \mathbb E [\log \hat p(y^{(0)}_i | y^{(1)}_j)] \\
&+ \sum_{\ell=1}^{L-1} \sum_{i=1}^{m_\ell} \sum_{j=1}^{m_{\ell+1}} \varepsilon^{(\ell)}_{ij} \mathbb E [\log \hat p(y_i^{(\ell)} | y_j^{(\ell+1)})] + \underset{q}{H} (y_i^{(\ell)} | \bm y^{(0)}) \\
&+ \sum_{i=1}^{m_L} p(y_i^{(L)}) + \underset{q}{H} (y_i^{(L)} | \bm y^{(0)})
\end{align*}

This objective separates as a sum of terms, each of which involves no more than one latent variable in the same layer. This allows any group of variables forming an independent set in the model graph to be updated in parallel, the same as for undirected models.

\newpage

\bibliographystyle{apalike}
\bibliography{uai_2018}

\end{document}